\documentclass[conference]{IEEEtran}

\usepackage{cite}
\usepackage{amsmath,amssymb,amsfonts}
\usepackage{graphicx}
\usepackage{booktabs}
\usepackage{multirow}
\usepackage{array}
\usepackage{tikz}
\usepackage{pgfplots}
\usepackage{url}
\usepackage{microtype}

\usetikzlibrary{arrows.meta,positioning,calc,fit}
\pgfplotsset{compat=1.18}

\newcommand{\pp}{\,\mathrm{pp}}
\newcommand{\sat}{\operatorname{sat}}
\newcommand{\clip}{\operatorname{clip}}

\begin{document}

\title{Bit-Accurate FPGA Evaluation of Learned Feature Gating in a Fixed-Point Fourier-Feature Automatic Modulation Classifier}

\author{
\IEEEauthorblockN{Gawthaman Senthilvelan}
\IEEEauthorblockA{
\textit{University of Toronto}\\
Toronto, Canada\\
gawthaman.senthilvelan@mail.utoronto.ca
}
\and
\IEEEauthorblockN{Luthira Abeykoon}
\IEEEauthorblockA{
\textit{University of Toronto}\\
Toronto, Canada\\
luthira.abeykoon@mail.utoronto.ca
}
}

\maketitle


\begin{abstract}
Learned feature reweighting can improve automatic modulation classification
(AMC) in software, but the same operation introduces additional arithmetic and
latency when implemented on an FPGA. This work measures that trade-off in a
compact fixed-point classifier using 24 sparse DFT-energy features, 8
phase/statistical features, and a 32-to-128-to-11 multilayer perceptron. A
second architecture inserts a learned 32-element, 8-bit input-dependent gate
before the classifier. Gated and ungated models are trained using
post-training quantization (PTQ) and quantization-aware training (QAT) with two
matched training seeds. The resulting 8 checkpoints are compiled independently
for an Intel Cyclone V FPGA and evaluated over 352,000 physical-board
classifications. Ungated models achieve higher test accuracy in all four
matched gate comparisons, with mean gated-minus-ungated differences of
$-0.784\pp$ under PTQ and $-0.616\pp$ under QAT. The effect of QAT changes
direction between the two training seeds. In hardware, the gate adds an average
of 1,318 ALMs, 1,557 registers, 4 DSP blocks, and 3,140 processing cycles.
All 352,000 board predictions agree exactly with an independent integer
reference, and 3,760 captured intermediate values from one training seed also
match. For this feature representation and implementation, learned gating
increases FPGA cost without improving classification accuracy.
\end{abstract}


\section{Introduction}

Automatic modulation classification (AMC) determines the modulation format of
an observed signal when that format is not known beforehand. Applications
include spectrum monitoring, cognitive radio, interference analysis, and
adaptive wireless receivers \cite{dobre2007}. Traditional AMC systems rely on
likelihood-based tests or engineered statistics, while more recent approaches
learn representations directly from in-phase and quadrature (I/Q) samples
\cite{oshea2016cnn,rajendran2018}.

Implementing these classifiers on an FPGA changes the design trade-offs.
Operations that are inexpensive in floating-point software may require
additional multipliers, registers, control states, or wider accumulators in
fixed-point hardware. FPGA AMC research has consequently explored quantized
RFSoC classifiers \cite{tridgell2020}, streaming inference architectures
\cite{jentzsch2022}, combined spectral and neural accelerators
\cite{jung2023}, ternary-weight networks \cite{woo2024}, and FPGA
implementations of attention-based models \cite{wang2025}.

Feature gating is particularly interesting in this setting. Attention and
gating mechanisms have improved several learned AMC models by allowing the
network to emphasize informative signal components
\cite{lin2022,chang2022,zhang2023}. In hardware, however, an input-dependent
gate requires its own transform, nonlinear mapping, and elementwise
multiplications. For a compact classifier, that additional computation can
represent a meaningful fraction of the complete accelerator.

This work evaluates whether that cost is worthwhile after the signal has
already been compressed into a small Fourier/statistical feature vector. The
experiment compares two classifiers that share the same 32-feature input,
128-neuron hidden layer, 11-class output layer, arithmetic widths, dataset
split, FPGA platform, and 50-MHz clock target. The only architectural
difference is the presence of a learned 32-element feature gate. PTQ and QAT
are evaluated separately, and the complete comparison is repeated with two
training seeds.

The main contributions are:

\begin{itemize}
    \item a fixed-point Cyclone-V AMC implementation using 24 sparse DFT-energy
    features and 8 phase/statistical features;
    \item a matched gated/ungated and PTQ/QAT comparison using 8 independently
    compiled FPGA bitstreams;
    \item physical-board evaluation over 352,000 classifications with exact
    board/reference prediction agreement; and
    \item paired accuracy, class-level, SNR, resource, and timing measurements
    that quantify the value and hardware cost of the learned gate.
\end{itemize}

The paired test set contains enough examples to measure differences between
specific checkpoints precisely. The two training seeds provide a separate,
smaller measure of whether those differences remain consistent after
retraining. Both views are reported because they answer different questions.

\section{Related Work}

\subsection{AMC Representations}

Classical AMC methods use features such as higher-order moments and cumulants,
instantaneous amplitude and phase, cyclostationary structure, and
frequency-domain statistics \cite{dobre2007}. Deep-learning approaches instead
learn representations from raw or transformed I/Q samples
\cite{oshea2016cnn,rajendran2018}.

RadioML 2016.10A was introduced through the GNU Radio machine-learning dataset
work \cite{oshea2016dataset} and remains a common controlled benchmark for AMC
research. The dataset is synthetic, so results on it primarily measure
performance under its simulated channel and signal-generation conditions.

The classifier used here combines engineered and learned processing. A
fixed signal-processing front end reduces each 128-sample I/Q window to 32
features, and an MLP performs the final classification. This compact interface
also makes it possible to measure the incremental hardware cost of an added
feature gate.

\subsection{Attention and Gating in AMC}

Several AMC architectures use attention or gating to adapt the representation
to the observed signal. Lin \emph{et al.} applied time-frequency attention to
modulation recognition \cite{lin2022}. Chang \emph{et al.} proposed a
convolutional gated network with a hierarchical classification head
\cite{chang2022}, and Zhang \emph{et al.} used higher-order attention to model
correlations among learned radio features \cite{zhang2023}.

These methods motivate adaptive feature weighting in AMC. The experiment in
this paper studies a different regime: the gate is inserted after a fixed
32-feature front end and must therefore justify its cost within a relatively
small FPGA classifier.

\subsection{Low-Precision and FPGA AMC}

Low-precision inference is widely used to reduce the implementation cost of
neural networks. Quantization-aware training can expose a model to
integer-like arithmetic during optimization and reduce the mismatch between
training and deployment \cite{jacob2018}.

For AMC, Tridgell \emph{et al.} demonstrated real-time quantized
classification on RFSoC \cite{tridgell2020}. Jentzsch \emph{et al.} mapped
RadioML models to FINN streaming architectures \cite{jentzsch2022}. Jung
\emph{et al.} combined an STFT front end with a fixed-point CNN accelerator
\cite{jung2023}, while Woo \emph{et al.} reduced implementation cost using
ternary weights \cite{woo2024}. Wang \emph{et al.} reported an FPGA
accelerator for a Transformer-based modulation recognizer \cite{wang2025}.

The present work complements these accelerator studies by focusing on an
operator-level ablation. Its objective is to measure how one learned
feature-processing stage changes classification accuracy, FPGA utilization,
and latency when the surrounding datapath is unchanged.

\section{Classifier and FPGA Architecture}

\subsection{System Boundary and Input}

Each observation contains $N=128$ complex samples,

\begin{equation}
    z[n]=I[n]+jQ[n], \qquad 0\le n<N.
    \label{eq:iq}
\end{equation}

Host software first performs per-window complex RMS normalization,

\begin{equation}
    r =
    \sqrt{
    \frac{1}{N}
    \sum_{n=0}^{N-1}
    \left(I[n]^2+Q[n]^2\right)+10^{-12}
    },
    \label{eq:rms}
\end{equation}

followed by

\begin{equation}
    I'[n]=\frac{I[n]}{r},
    \qquad
    Q'[n]=\frac{Q[n]}{r}.
\end{equation}

The normalized values are rounded to nearest, clipped, and converted to signed
Q1.15 before transfer through Virtual JTAG. FPGA processing therefore begins
with normalized fixed-point I/Q samples.


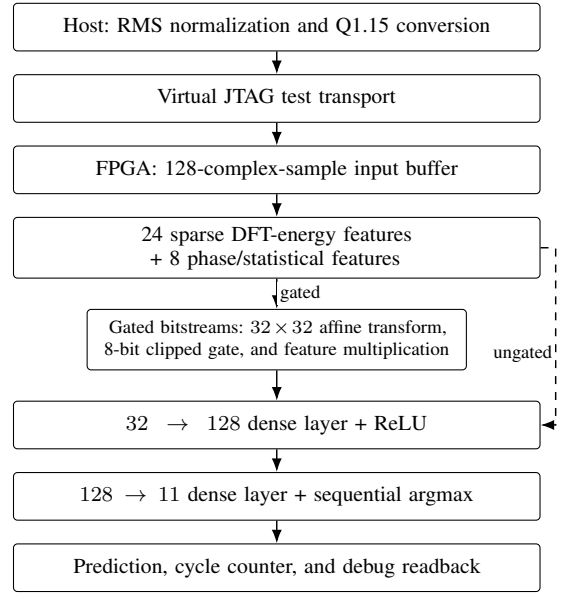
\begin{figure}[t]
\centering
\begin{tikzpicture}[
    node distance=3.0mm,
    block/.style={
        draw,
        rounded corners=1.5pt,
        align=center,
        text width=0.76\columnwidth,
        minimum height=6.3mm,
        font=\footnotesize
    },
    gateblock/.style={
        draw,
        rounded corners=1.5pt,
        align=center,
        text width=0.54\columnwidth,
        minimum height=8mm,
        font=\scriptsize
    },
    flow/.style={
        -{Latex[length=1.8mm]},
        line width=0.55pt
    },
    bypass/.style={
        -{Latex[length=1.8mm]},
        dashed,
        line width=0.55pt
    },
    lbl/.style={
        font=\scriptsize,
        fill=white,
        inner sep=1pt
    }
]

\node[block] (host)
{Host: RMS normalization and Q1.15 conversion};

\node[block, below=of host] (jtag)
{Virtual JTAG test transport};

\node[block, below=of jtag] (buf)
{FPGA: 128-complex-sample input buffer};

\node[block, below=of buf] (feat)
{24 sparse DFT-energy features + 8 phase/statistical features};

\node[gateblock, below=4mm of feat] (gate)
{Gated bitstreams: $32\!\times\!32$ affine transform,\\
8-bit clipped gate, and feature multiplication};

\node[block, below=4mm of gate] (hidden)
{$32\!\rightarrow\!128$ dense layer + ReLU};

\node[block, below=of hidden] (out)
{$128\!\rightarrow\!11$ dense layer + sequential argmax};

\node[block, below=of out] (read)
{Prediction, cycle counter, and debug readback};

\draw[flow] (host) -- (jtag);
\draw[flow] (jtag) -- (buf);
\draw[flow] (buf) -- (feat);

\draw[flow]
(feat) --
node[right,lbl]{gated}
(gate);

\draw[flow] (gate) -- (hidden);

\draw[bypass]
(feat.east) --
++(2mm,0)
|- node[pos=0.30,left,lbl]{ungated}
(hidden.east);

\draw[flow] (hidden) -- (out);
\draw[flow] (out) -- (read);

\end{tikzpicture}

\caption{Experimental pipeline. Host processing performs RMS normalization and
Q1.15 conversion. Both FPGA variants use the same feature extractor and
classifier; the ungated variant bypasses the learned feature gate.}
\label{fig:arch}
\end{figure}

\subsection{Sparse Fourier Features}

The first 24 features sample approximately uniform bins of a 128-point DFT:

\begin{equation}
\begin{split}
\mathcal{K}=\{&
0,5,11,16,21,27,32,37,43,48,53,59,\\
&64,69,75,80,85,91,96,101,107,112,117,123
\}.
\end{split}
\label{eq:bins}
\end{equation}

For bin $k_j$, signed Q1.15 cosine and sine coefficients $c_j[n]$ and $s_j[n]$
are used to compute

\begin{align}
R_j
&=
\sum_n
I'[n]c_j[n]+Q'[n]s_j[n],
\\
J_j
&=
\sum_n
Q'[n]c_j[n]-I'[n]s_j[n].
\end{align}

Both use 48-bit accumulators. The corresponding energy feature is

\begin{equation}
f_j=
\sat_{32}
\left(
(R_j^2+J_j^2)\gg56
\right).
\label{eq:dftfeature}
\end{equation}

The sparse set provides coarse frequency-domain energy information while
avoiding the cost of computing and retaining all 128 DFT bins.

\subsection{Phase and Statistical Features}

Eight additional features retain signal information that is absent from the
DFT magnitudes. Defining

\begin{equation}
z_n=I'[n]+jQ'[n],
\end{equation}

the features are

\begin{align}
&\Re\left(\sum_n z_n^2\right),
\qquad
\Im\left(\sum_n z_n^2\right),
\\
&\Re\left(\sum_n z_n^4\right),
\qquad
\Im\left(\sum_n z_n^4\right),
\\
&\Re\left(
\sum_{n=1}^{N-1}
z_nz_{n-1}^{*}
\right),
\qquad
\Im\left(
\sum_{n=1}^{N-1}
z_nz_{n-1}^{*}
\right),
\\
&\sum_n |z_n|^4,
\qquad
\sum_n I'[n]Q'[n].
\end{align}

Second-order terms are shifted by 24 bits and fourth-order terms by 54 bits
before 32-bit saturation. Together with the 24 spectral features, these values
form

\begin{equation}
\mathbf{x}\in\mathbb{Z}^{32}.
\end{equation}

\subsection{Gated and Ungated Classifiers}

The gated classifier first applies a learned 32-to-32 affine transform,

\begin{align}
\mathbf{p}
&=
(W_g\mathbf{x}+\mathbf{b}_g)\gg26,
\\
\mathbf{g}
&=
\clip(\mathbf{p}+128,0,255),
\\
\mathbf{x}_g
&=
(\mathbf{x}\odot\mathbf{g})\gg8.
\label{eq:gate}
\end{align}

The 8-bit gate therefore performs continuous feature attenuation. It is not a
binary feature selector. In the ungated architecture,

\begin{equation}
\mathbf{x}_g=\mathbf{x}.
\end{equation}

Both variants then evaluate

\begin{align}
\mathbf{h}
&=
\operatorname{ReLU}
(W_1\mathbf{x}_g+\mathbf{b}_1),
\\
\mathbf{y}
&=
W_2\mathbf{h}+\mathbf{b}_2,
\\
\hat{c}
&=
\arg\max_c y_c.
\end{align}

The gated model contains 6,699 learned parameters, compared with 5,643 for the
ungated model. This is an increase of 1,056 parameters, or 18.7\%.
Approximate parameter storage under the deployed widths is 13.74~kB for the
gated model and 11.56~kB for the ungated model.

Debug captures from four seed-20260716 test windows produced gate values from
122 to 192, with an average of approximately

\begin{equation}
142/256=0.555.
\end{equation}

None of these captured values reached either clipping boundary.

\subsection{Sequential Hardware Schedule}

The implementation reuses arithmetic units to limit resource consumption.
DFT bins are processed sequentially, and both dense layers reuse
multiply-accumulate hardware. The gated architecture also computes gate
outputs and gated features sequentially. A final sequential comparison selects
the maximum logit.

The complete gated datapath requires 27,374 core cycles per observation,
compared with 24,234 cycles for the ungated datapath. The 3,140 added cycles
come from the $32\times32$ gate transform, 32 gate-feature products, and their
controller states.

\section{Fixed-Point Co-Design}

Table~\ref{tab:fixed} summarizes the deployed arithmetic. Parameter conversion
uses round-to-nearest followed by signed clipping, while right shifts in the
RTL perform arithmetic truncation. No hidden-layer or output-layer saturation
events were observed in the retained validation and test evaluations.

\begin{table}[t]
\caption{Deployed Fixed-Point Contract}
\label{tab:fixed}
\centering
\scriptsize

\begin{tabular}{p{0.43\columnwidth}p{0.47\columnwidth}}
\toprule
\textbf{Quantity} &
\textbf{Representation / scaling}
\\
\midrule

I/Q samples &
signed Q1.15
\\

DFT coefficients &
signed Q1.15
\\

DFT accumulators &
signed 48-bit
\\

Extracted features &
signed 32-bit; effective Q16 interface
\\

Gate weights / biases &
int16 $\times2^{10}$ / int32 $\times2^{26}$
\\

Gate output &
unsigned 8-bit
\\

Dense-1 weights / biases &
int16 $\times2^{10}$ / int32 $\times2^{26}$
\\

Dense-1 output shift &
18 bits; hidden fractional bits = 8
\\

Dense-2 weights / biases &
int16 $\times2^{8}$ / int32 $\times2^{16}$
\\

Dense-2 output shift &
16 bits
\\

Dense accumulators &
signed 64-bit
\\

Logits &
signed 32-bit integer
\\

\bottomrule
\end{tabular}
\end{table}

An independent integer implementation reproduces the deployed coefficient
files, word lengths, shifts, clipping, ReLU operation, and argmax. It is used
during checkpoint evaluation and again after FPGA deployment to verify the
physical-board outputs.

\section{Experimental Methodology}

\subsection{Dataset and Frozen Split}

RadioML 2016.10A contains 220,000 signal windows representing 11 modulation
classes at 20 SNR values between $-20$ and $+18$~dB in 2-dB increments
\cite{oshea2016dataset}. Each class--SNR combination contains 1,000 windows.

A class/SNR-stratified permutation generated using split seed 20260716 is used
throughout the study. For each class--SNR combination, 600 samples are assigned
to training, 200 to validation, and 200 to testing. This produces 132,000
training examples, 44,000 validation examples, and 44,000 test examples.

All eight FPGA evaluations use the same ordered set of 44,000 test windows.

\begin{table}[t]
\caption{Dataset Partition}
\label{tab:split}
\centering

\begin{tabular}{lrr}
\toprule
\textbf{Partition} &
\textbf{Per class--SNR key} &
\textbf{Total}
\\
\midrule

Training &
600 &
132,000
\\

Validation &
200 &
44,000
\\

Test &
200 &
44,000
\\

\bottomrule
\end{tabular}
\end{table}

\subsection{Matched Training Design}

Two training seeds are evaluated:

\begin{equation}
20260716
\qquad\text{and}\qquad
20260719.
\end{equation}

Within each seed, gated and ungated models use identical initialization for the
shared dense layers. The gated network additionally initializes its gate.
Training-data ordering is controlled by the same seed.

Base training runs for 45 epochs using Adam with:

\begin{itemize}
    \item batch size: 2,048;
    \item learning rate: 0.0025;
    \item $\beta_1=0.9$;
    \item $\beta_2=0.999$;
    \item $\epsilon=10^{-8}$; and
    \item multiclass cross entropy.
\end{itemize}

For PTQ, training is performed in floating point. At the end of each epoch, the
model is converted using the deployed fixed-point arithmetic and evaluated on
the validation set. The epoch with the highest integer validation accuracy is
selected.

QAT begins from the selected base checkpoint and continues for 15 epochs with
a learning rate of 0.0005. The forward pass includes fake-quantized weights and
biases together with the deployed gate, gated-feature, hidden-layer, and
output-layer truncations. A straight-through estimator is used through the
unsaturated gate range.

QAT checkpoints are also selected using exact integer validation accuracy.
The test set is used only after checkpoint selection.

\begin{table*}[t]
\caption{Selected Checkpoints from Exact Fixed-Point Validation}
\label{tab:checkpoints}
\centering

\begin{tabular}{lcc|lcc}
\toprule

\textbf{Checkpoint} &
\textbf{Epoch} &
\textbf{Val. acc.} &
\textbf{Checkpoint} &
\textbf{Epoch} &
\textbf{Val. acc.}
\\

\midrule

Gated PTQ, 20260716 &
45 &
46.289\% &
Gated PTQ, 20260719 &
45 &
46.570\%
\\

Gated QAT, 20260716 &
8 &
46.720\% &
Gated QAT, 20260719 &
9 &
46.436\%
\\

Ungated PTQ, 20260716 &
42 &
47.111\% &
Ungated PTQ, 20260719 &
45 &
47.375\%
\\

Ungated QAT, 20260716 &
15 &
47.250\% &
Ungated QAT, 20260719 &
11 &
47.205\%
\\

\bottomrule
\end{tabular}
\end{table*}

\subsection{Physical-Board Evaluation}

Each selected checkpoint is exported to fixed-point memory files and compiled
into a separate Quartus bitstream. The bitstreams are programmed onto a
Terasic DE1-SoC containing an Intel Cyclone V 5CSEMA5F31C6 FPGA.

Quartus Prime Lite 18.1 Build 625 is used with a 50-MHz clock target.

A 4,096-bit Virtual-JTAG payload transfers each normalized I/Q window to the
FPGA. The status interface returns the predicted class, cumulative correctness
and test counters, the core-cycle count, and selected debug values.

Latency is measured from the start of FPGA computation to completion of the
class decision. JTAG transfer time is excluded. Transfer, buffering, and
inference are not overlapped in the current prototype, so application-level
streaming throughput is not characterized.

Power and energy measurements were not collected.

\subsection{Verification and Statistical Analysis}

The prediction from every physical-board classification is compared against
the independent integer implementation.

For seed 20260716, four test observations from each checkpoint are also
captured at intermediate stages. These captures include extracted features,
gate and gated-feature values where applicable, hidden activations, and logits.

A separate 17-vector ModelSim suite exercises arithmetic corner cases in the
RTL.

Accuracy intervals are calculated using 95\% Wilson intervals. Gated and
ungated checkpoints within a seed process identical test observations, allowing
their prediction differences to be evaluated with paired McNemar counts.

These paired tests describe the behavior of individual trained checkpoints.
Variation between independently trained models is summarized separately using
the two training seeds.

\section{Results}

\subsection{Physical-Board Accuracy}

Table~\ref{tab:accuracy} reports the eight FPGA evaluations. The highest
individual accuracy is 47.257\%, obtained by the seed-20260719 ungated PTQ
model.

For the primary architecture comparison, the ungated classifier exceeds the
gated classifier for both training seeds under both PTQ and QAT.

\begin{table*}[t]
\caption{Physical-Board Accuracy on the Same 44,000 Test Windows per Run}
\label{tab:accuracy}
\centering

\begin{tabular}{llrcc}
\toprule

\textbf{Architecture} &
\textbf{Quantization} &
\textbf{Seed} &
\textbf{Correct / 44,000} &
\textbf{Accuracy [Wilson 95\% CI]}
\\

\midrule

Gated &
PTQ &
20260716 &
20,223 &
45.961\% [45.496, 46.427]
\\

Gated &
PTQ &
20260719 &
20,485 &
46.557\% [46.091, 47.023]
\\

Gated &
QAT &
20260716 &
20,394 &
46.350\% [45.884, 46.816]
\\

Gated &
QAT &
20260719 &
20,364 &
46.282\% [45.816, 46.748]
\\

Ungated &
PTQ &
20260716 &
20,605 &
46.830\% [46.364, 47.296]
\\

Ungated &
PTQ &
20260719 &
20,793 &
\textbf{47.257\%} [46.791, 47.724]
\\

Ungated &
QAT &
20260716 &
20,661 &
46.957\% [46.491, 47.423]
\\

Ungated &
QAT &
20260719 &
20,639 &
46.907\% [46.441, 47.373]
\\

\bottomrule
\end{tabular}
\end{table*}

The two-seed mean accuracies are

\begin{align}
\text{Gated PTQ} &= 46.259\%,\\
\text{Gated QAT} &= 46.316\%,\\
\text{Ungated PTQ} &= 47.043\%,\\
\text{Ungated QAT} &= 46.932\%.
\end{align}

The corresponding seed-to-seed standard deviations are 0.421, 0.048, 0.302,
and 0.035 percentage points.

\subsection{Gate and QAT Contrasts}

Table~\ref{tab:contrasts} gives the matched architecture and quantization
contrasts.

Under PTQ, the gated-minus-ungated accuracy differences are

\begin{equation}
-0.868\pp
\qquad\text{and}\qquad
-0.700\pp
\end{equation}

for seeds 20260716 and 20260719, respectively. Under QAT, the differences are

\begin{equation}
-0.607\pp
\qquad\text{and}\qquad
-0.625\pp.
\end{equation}

Their two-seed means are therefore $-0.784\pp$ for PTQ and
$-0.616\pp$ for QAT. The paired prediction counts in Table~\ref{tab:contrasts}
show the same direction in each comparison.

The QAT results are less consistent. Relative to PTQ, QAT changes gated
accuracy by $+0.389\pp$ for seed 20260716 and $-0.275\pp$ for seed 20260719.
For the ungated classifier, the corresponding changes are $+0.127\pp$ and
$-0.350\pp$. The available replications therefore show no consistent
accuracy improvement from QAT.

\begin{table*}[t]
\caption{Matched Accuracy Contrasts. ``First'' and ``second'' follow the order
named in each contrast. McNemar counts use identical test windows within each
training seed.}
\label{tab:contrasts}
\centering
\scriptsize

\begin{tabular}{llrrrcc}
\toprule

\textbf{Contrast} &
\textbf{Seed} &
\textbf{$\Delta$ acc.} &
\textbf{First-only correct} &
\textbf{Second-only correct} &
\textbf{$p$} &
\textbf{Two-seed mean}
\\

\midrule

Gated $-$ ungated, PTQ &
20260716 &
$-0.868\pp$ &
1,281 &
1,663 &
$2.04\times10^{-12}$ &
\multirow{2}{*}{$-0.784\pp$}
\\

&
20260719 &
$-0.700\pp$ &
1,722 &
2,030 &
$5.31\times10^{-7}$ &
\\

\addlinespace

Gated $-$ ungated, QAT &
20260716 &
$-0.607\pp$ &
1,511 &
1,778 &
$3.47\times10^{-6}$ &
\multirow{2}{*}{$-0.616\pp$}
\\

&
20260719 &
$-0.625\pp$ &
1,782 &
2,057 &
$9.69\times10^{-6}$ &
\\

\addlinespace

QAT $-$ PTQ, gated &
20260716 &
$+0.389\pp$ &
-- &
-- &
-- &
\multirow{2}{*}{$+0.057\pp$}
\\

&
20260719 &
$-0.275\pp$ &
-- &
-- &
-- &
\\

QAT $-$ PTQ, ungated &
20260716 &
$+0.127\pp$ &
-- &
-- &
-- &
\multirow{2}{*}{$-0.112\pp$}
\\

&
20260719 &
$-0.350\pp$ &
-- &
-- &
-- &
\\

\bottomrule
\end{tabular}
\end{table*}

\subsection{Accuracy Versus SNR}

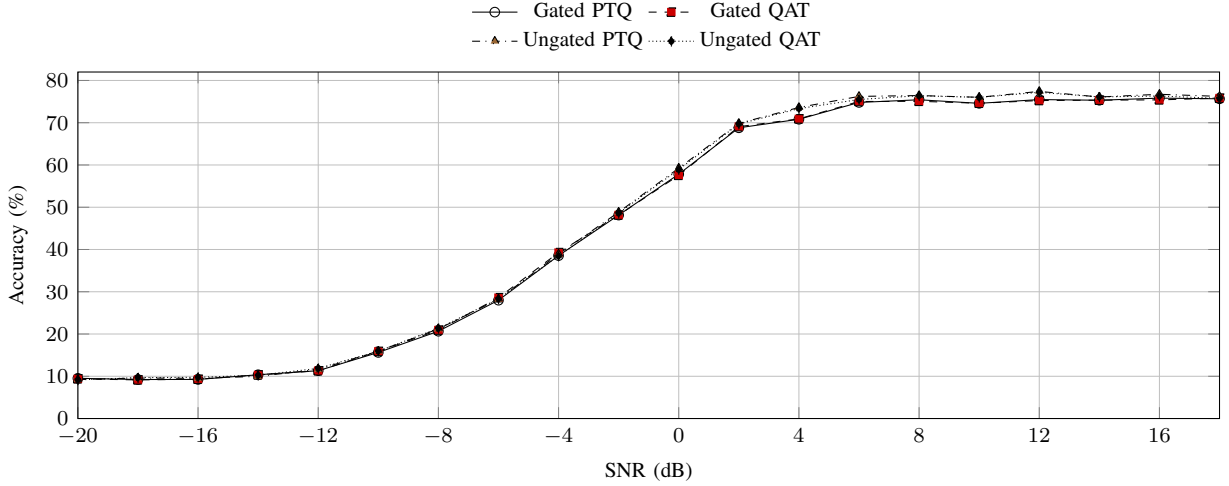
\begin{figure*}[t]
\centering

\begin{tikzpicture}

\begin{axis}[
    width=0.92\textwidth,
    height=0.34\textwidth,
    xlabel={SNR (dB)},
    ylabel={Accuracy (\%)},
    xmin=-20,
    xmax=18,
    ymin=0,
    ymax=82,
    xtick={-20,-16,-12,-8,-4,0,4,8,12,16},
    ytick={0,10,20,30,40,50,60,70,80},
    grid=both,
    major grid style={line width=.2pt},
    legend columns=2,
    legend style={
        at={(0.5,1.03)},
        anchor=south,
        draw=none,
        font=\footnotesize
    },
    tick label style={font=\footnotesize},
    label style={font=\footnotesize}
]

\addplot+[
    black,
    solid,
    mark=o,
    mark size=1.8pt
]
coordinates {
(-20,9.52)
(-18,9.23)
(-16,9.25)
(-14,10.36)
(-12,11.32)
(-10,15.64)
(-8,20.66)
(-6,28.00)
(-4,38.55)
(-2,48.14)
(0,57.82)
(2,68.80)
(4,70.82)
(6,74.80)
(8,75.43)
(10,74.59)
(12,75.48)
(14,75.32)
(16,75.82)
(18,75.66)
};
\addlegendentry{Gated PTQ}

\addplot+[
    black,
    dashed,
    mark=square*,
    mark size=1.6pt
]
coordinates {
(-20,9.39)
(-18,9.14)
(-16,9.30)
(-14,10.32)
(-12,11.30)
(-10,15.89)
(-8,20.93)
(-6,28.59)
(-4,39.27)
(-2,48.16)
(0,57.55)
(2,69.07)
(4,70.91)
(6,75.02)
(8,75.07)
(10,74.61)
(12,75.25)
(14,75.34)
(16,75.41)
(18,75.82)
};
\addlegendentry{Gated QAT}

\addplot+[
    black,
    dashdotted,
    mark=triangle*,
    mark size=1.8pt
]
coordinates {
(-20,9.20)
(-18,9.59)
(-16,9.50)
(-14,9.98)
(-12,11.61)
(-10,15.95)
(-8,21.20)
(-6,28.23)
(-4,38.91)
(-2,48.84)
(0,59.25)
(2,69.84)
(4,73.61)
(6,76.23)
(8,76.43)
(10,76.02)
(12,77.41)
(14,76.09)
(16,76.73)
(18,76.23)
};
\addlegendentry{Ungated PTQ}

\addplot+[
    black,
    densely dotted,
    mark=diamond*,
    mark size=1.8pt
]
coordinates {
(-20,9.14)
(-18,9.68)
(-16,9.75)
(-14,10.25)
(-12,11.86)
(-10,15.95)
(-8,21.34)
(-6,28.32)
(-4,38.45)
(-2,48.61)
(0,58.93)
(2,69.70)
(4,73.34)
(6,75.50)
(8,76.39)
(10,76.00)
(12,77.20)
(14,76.11)
(16,76.32)
(18,75.77)
};
\addlegendentry{Ungated QAT}

\end{axis}
\end{tikzpicture}

\caption{Mean physical-board accuracy versus SNR across the two training
seeds.}
\label{fig:snr}
\end{figure*}

Figure~\ref{fig:snr} shows mean physical-board accuracy across the two training
seeds at each SNR.

Accuracy remains close to the 11-class chance level,

\begin{equation}
\frac{1}{11}=9.09\%,
\end{equation}

at the lowest SNRs. It then increases rapidly through the middle of the SNR
range before reaching approximately 75--77\% at high SNR.

The separation between gated and ungated models is most visible from moderate
to high SNR, where additive noise is less dominant.

Across broader SNR bands, gated and ungated PTQ accuracy is 20.07/20.30\% from
$-20$ to $-2$~dB, 69.53/71.07\% from 0 to 8~dB, and 75.37/76.50\% from 10
to 18~dB.

For QAT, the corresponding values are 20.23/20.34\%, 69.52/70.77\%, and
75.29/76.28\%.

\subsection{Class-Level Behavior}

Table~\ref{tab:classrecall} reports mean recall for each modulation class
across the two training seeds.

AM-SSB achieves the highest recall in all four model conditions. QAM64, WBFM,
and QAM16 are substantially weaker. Their pooled high-SNR recalls are 18.8\%,
32.2\%, and 57.2\%, respectively, showing that these errors remain even after
the low-SNR portion of the dataset is excluded.

\begin{table*}[t]
\caption{Mean Per-Class Recall Across the Two Physical-Board Seeds}
\label{tab:classrecall}
\centering
\scriptsize

\begin{tabular}{lrrrrrrrrrrr}
\toprule

&
\textbf{8PSK} &
\textbf{AM-DSB} &
\textbf{AM-SSB} &
\textbf{BPSK} &
\textbf{CPFSK} &
\textbf{GFSK} &
\textbf{PAM4} &
\textbf{QAM16} &
\textbf{QAM64} &
\textbf{QPSK} &
\textbf{WBFM}
\\

\midrule

Gated PTQ &
53.63 &
71.89 &
83.85 &
54.34 &
42.19 &
50.19 &
51.50 &
28.78 &
9.31 &
44.04 &
19.15
\\

Gated QAT &
54.36 &
70.94 &
83.86 &
55.39 &
41.94 &
51.16 &
50.74 &
29.08 &
7.49 &
44.46 &
20.06
\\

Ungated PTQ &
53.49 &
72.69 &
85.63 &
55.33 &
44.06 &
50.78 &
51.83 &
29.59 &
9.31 &
45.63 &
19.16
\\

Ungated QAT &
57.53 &
70.91 &
81.09 &
56.09 &
43.00 &
51.68 &
51.84 &
28.05 &
9.45 &
45.61 &
21.01
\\

\bottomrule
\end{tabular}
\end{table*}

The pooled predictions also reveal a strong class imbalance in the classifier
outputs. AM-SSB accounts for 31.20\% of all predictions even though each true
class contributes 9.09\% of the test set. QAM64 and WBFM account for only
1.58\% and 2.43\% of predictions.

Table~\ref{tab:confusions} lists the largest pooled confusion pairs.

\begin{table}[t]
\caption{Dominant Pooled Confusions Across All 8 Board Runs}
\label{tab:confusions}
\centering
\scriptsize

\begin{tabular}{lr}
\toprule
\textbf{True $\rightarrow$ predicted} &
\textbf{Rate}
\\
\midrule

WBFM $\rightarrow$ AM-DSB &
43.61\%
\\

QAM64 $\rightarrow$ QAM16 &
30.73\%
\\

BPSK $\rightarrow$ AM-SSB &
28.22\%
\\

8PSK $\rightarrow$ AM-SSB &
27.82\%
\\

QPSK $\rightarrow$ AM-SSB &
27.66\%
\\

PAM4 $\rightarrow$ AM-SSB &
27.63\%
\\

QAM16 $\rightarrow$ AM-SSB &
27.32\%
\\

CPFSK $\rightarrow$ AM-SSB &
27.15\%
\\

QAM16 $\rightarrow$ 8PSK &
25.72\%
\\

QAM64 $\rightarrow$ 8PSK &
23.41\%
\\

\bottomrule
\end{tabular}
\end{table}

One possible cause is the compact feature representation. The front end
captures coarse spectral energy and low-order complex statistics, but does not
explicitly model long FM trajectories, detailed amplitude distributions,
symbol timing, or local constellation structure. The QAM16/QAM64 and
WBFM/AM-DSB confusion patterns are consistent with this explanation, although
targeted feature ablations would be needed to establish it directly.

\subsection{FPGA Resource and Timing Cost}

Table~\ref{tab:hardware} summarizes the four compiled builds for each
architecture. ALM and register utilization vary slightly with checkpoint
parameters because constant values affect synthesis and logic packing. DSP use
and cycle count are identical across checkpoints of the same architecture.

\begin{table*}[t]
\caption{Cyclone-V Implementation Cost Across 4 Builds per Architecture}
\label{tab:hardware}
\centering

\begin{tabular}{lccc}
\toprule

\textbf{Metric} &
\textbf{Gated} &
\textbf{Ungated} &
\textbf{Gate overhead}
\\

\midrule

ALMs, mean (range) &
13,651 (13,597--13,712) &
12,334 (12,288--12,354) &
+1,318 / +10.7\%
\\

Approx. ALM utilization &
43\% &
39\% &
+4 percentage points
\\

Registers, mean &
22,745 &
21,188 &
+1,557 / +7.3\%
\\

DSP blocks &
21 / 87 &
17 / 87 &
+4 / +23.5\% relative
\\

DSP utilization &
24\% &
20\% &
+4 percentage points
\\

M10K blocks / bits &
0 / 0 &
0 / 0 &
0
\\

Core cycles &
27,374 &
24,234 &
+3,140 / +12.96\%
\\

Core latency at 50 MHz &
547.48~$\mu$s &
484.68~$\mu$s &
+62.80~$\mu$s
\\

Slow 85$^\circ$C Fmax range &
60.44--62.69 MHz &
60.17--62.08 MHz &
no material shift
\\

Minimum 50-MHz setup slack &
3.454 ns &
3.381 ns &
both close timing
\\

\bottomrule
\end{tabular}
\end{table*}

DSP hierarchy identifies the source of the four additional blocks in the gated
architecture. Two DSP blocks are used by the gate's dense transform and two
by feature multiplication. In both designs, the DFT engine uses 13 DSP blocks,
Dense-1 uses 2, and Dense-2 uses 2.

The additional gate has little effect on maximum clock frequency. Its
implementation cost is instead visible in arithmetic resources and processing
cycles.

Quartus maps neither the learned parameters nor the I/Q buffer into M10K
memory. Constants are implemented using synthesized logic and constant
structures, while the I/Q buffer is register based.

\subsection{Bit-Exact Deployment Verification}

Across all eight bitstreams, every physical-board class decision agrees with
the independent integer implementation:

\begin{equation}
352,000/352,000.
\end{equation}

For the four seed-20260716 checkpoints, the debug interface also captured
3,760 intermediate values from the feature, gate/gated-feature, hidden, and
logit stages. All 3,760 agree with the corresponding integer values.

\begin{table}[t]
\caption{Deployment-Equivalence Evidence}
\label{tab:verify}
\centering
\scriptsize

\begin{tabular}{p{0.58\columnwidth}r}
\toprule

\textbf{Check} &
\textbf{Result}
\\

\midrule

Final board/reference predictions, 8 runs &
352,000 / 352,000
\\

Intermediate board/reference values, seed 20260716 &
3,760 / 3,760
\\

Final prediction mismatches &
0
\\

Intermediate mismatches in retained captures &
0
\\

17-vector ModelSim arithmetic stress suite &
pass; 0 errors/warnings
\\

\bottomrule
\end{tabular}
\end{table}

The agreement confirms that the FPGA executes the intended fixed-point
checkpoints for the measured test cases.

\section{Discussion}

\subsection{Effect of Learned Gating}

The four matched architecture comparisons give the same outcome: inserting the
gate reduces test accuracy while increasing implementation cost.

A likely explanation is that the gate operates on an already compressed
32-dimensional representation. The following 32-to-128 dense layer has an
independent weight for each input-feature/hidden-neuron pair and can therefore
learn feature-dependent scaling itself. The extra 32-to-32 transform may add
little useful representational capacity in this setting.

The captured gate values provide some supporting evidence. Values from the
seed-20260716 debug samples remain between 122 and 192 and do not approach
either clipping boundary. In those samples, the gate behaves primarily as a
moderate rescaling stage rather than a strong selector.

Its fixed-point implementation nevertheless requires another affine transform,
8-bit clipping, and 32 feature multiplications. These operations increase DSP
use from 17 to 21 blocks and increase latency by 3,140 cycles.

The result applies to the gate studied here. Different conclusions may arise
with a larger feature representation, lower-rank gating, different gate
precision, or attention applied directly to temporal I/Q features.

\subsection{QAT and Representation Limits}

QAT does not produce a consistent gain across the two training seeds. It
improves the gated and ungated seed-20260716 models slightly, but reduces
accuracy for both seed-20260719 models.

The overall error pattern also suggests that quantization is not the main
accuracy bottleneck. At high SNR, all four conditions plateau near 75--77\%.
QAM64 and WBFM remain difficult, and AM-SSB is substantially overpredicted.

These errors point toward limitations in the 32-feature front end. Coarse
spectral energy and low-order statistics distinguish several modulation
families well, but they omit some temporal and amplitude structure useful for
separating the weakest classes. Improving the representation may therefore
have a larger effect than adding another reweighting stage to the existing
features.

\section{Limitations and Reproducibility}

Only two independently trained seeds are included. The gate comparison has the
same direction in both seeds and both quantization conditions, but additional
training replications would provide a better estimate of architecture-level
variance.

RadioML 2016.10A is a synthetic benchmark. This study does not include captured
RF signals, an external AMC dataset, or channel-distribution shifts. The host
also performs RMS normalization before the FPGA, so absolute received-power
information is absent from the hardware input.

The implementation is a board-level classifier prototype rather than a
complete streaming receiver. Test vectors are delivered through Virtual JTAG,
and inference is not overlapped with transport. The reported latency therefore
describes the classifier core. Power and energy were not measured.

The hardware was designed to compare two architectures under the same
conditions rather than to maximize throughput or minimize area. In particular,
the current design uses sequential scheduling, register-based buffering, and
no M10K blocks.

The retained experimental artifacts include the eight board-result CSV files,
integer-reference predictions, bitstreams, Quartus reports, checkpoints,
manifests, and the seed-20260716 intermediate captures. The exact frozen
\texttt{board\_test\_payloads.hex} and associated feature-cache file used
during the campaign are currently absent from the retained checkout. The board
runs are linked to a recorded payload hash, but the exact 44,000-window
fixed-point test payload should be restored before archival or submission to
enable complete reproduction of the experiment.

\section{Conclusion}

This work measured the software and hardware effect of adding an
input-dependent feature gate to a compact fixed-point AMC classifier. Eight
independently compiled FPGA checkpoints were evaluated across two training
seeds and two quantization procedures.

Ungated models achieve higher accuracy in all four matched architecture
comparisons. Their mean advantage is $0.784\pp$ under PTQ and $0.616\pp$
under QAT. Removing the gate also saves 4 DSP blocks, approximately 1.3k ALMs,
1.6k registers, and 3,140 core cycles.

QAT produces smaller and seed-dependent changes, with no common direction
across the two replications. All 352,000 physical-board predictions match the
integer reference exactly.

For this 32-feature Fourier/statistical classifier, the learned gate adds
hardware complexity without improving classification performance. The result
also highlights the importance of evaluating learned signal-processing
operators after fixed-point mapping, where their implementation cost becomes
part of the model-design decision.


\end{document}